\documentclass[a4paper]{article}

\usepackage{authblk}
\usepackage[utf8]{inputenc}
\usepackage[T1]{fontenc}
\usepackage{pslatex}
\usepackage[english]{babel}
\usepackage{fancyhdr}
\usepackage{hyperref}
\usepackage{amsmath,amssymb,amsfonts,amsthm}
\usepackage{graphicx}
\usepackage{amsmath}
\usepackage{amsthm}
\usepackage{amsfonts}
\usepackage{mathtools}
\usepackage{amssymb}
\usepackage{algorithm}
\usepackage{algorithmic}
\usepackage{microtype}
\usepackage{rotating}
\usepackage{todonotes}
\usepackage{tikz}
\usepackage{tikz-qtree}
\usepackage{xparse}
\usepackage{appendix}
\usepackage{url}
\usepackage{tabularx}
\usepackage{paralist}
\usepackage{booktabs}
\usepackage{multirow}
\usepackage{subcaption}
\usepackage{cancel}
\usetikzlibrary{calc,shapes.callouts,shapes.arrows}
\usetikzlibrary{automata}
\usetikzlibrary{positioning}
\usetikzlibrary{decorations.pathreplacing}
\usetikzlibrary{decorations.pathmorphing}
\usetikzlibrary{arrows}
\usetikzlibrary{shapes}

\newcommand{\N}{\mathbb{N}}
\newcommand{\Z}{\mathbb{Z}}
\newcommand{\F}{\mathbb{F}}

\newtheorem{definition}{Definition}

\newtheorem*{problem*}{Problem}

\providecommand{\keywords}[1]{\textbf{\textit{Keywords }} #1}

\begin{document}

\title{A Discrete Particle Swarm Optimizer for the Design of Cryptographic Boolean Functions\footnote{This manuscript is an extended version of a poster paper published in GECCO 2015~\cite{Mariot15}}}

\author[1]{Luca Mariot}
\author[2]{Alberto Leporati}
\author[3]{Luca Manzoni}

\affil[1]{{\normalsize Semantics, Cybersecurity and Services Group, University of Twente, 7522 NB Enschede, The Netherlands} \\
	
	{\small \texttt{l.mariot@utwente.nl}}}

\affil[3]{{\normalsize Dipartimento di Informatica, Sistemistica e Comunicazione, Università degli Studi di Milano-Bicocca, Viale Sarca 336/14, 20126, Milano, Italy} \\
	
	{\small \texttt{lmanzoni@units.it}}}

\affil[3]{{\normalsize Dipartimento di Matematica, Informatica e Geoscienze, Università degli Studi di Trieste, Via Alfonso Valerio 12/1, Trieste, 34127, Italy } \\
	
	{\small \texttt{lmanzoni@units.it}}}
	
\maketitle

\begin{abstract}
A Particle Swarm Optimizer for the search of balanced Boolean functions with good cryptographic properties is proposed in this paper. The algorithm is a modified version of the permutation PSO by Hu, Eberhart and Shi which preserves the Hamming weight of the particles positions, coupled with the Hill Climbing method devised by Millan, Clark and Dawson to improve the nonlinearity and deviation from correlation immunity of Boolean functions. The parameters for the PSO velocity equation are tuned by means of two meta-optimization techniques, namely Local Unimodal Sampling (LUS) and Continuous Genetic Algorithms (CGA), finding that CGA produces better results. Using the CGA-evolved parameters, the PSO algorithm is then run on the spaces of Boolean functions from $n=7$ to
$n=12$ variables. The results of the experiments are reported, observing that this new PSO algorithm generates Boolean functions featuring similar or better combinations of nonlinearity, correlation immunity and propagation criterion with respect to the ones obtained by other optimization methods.
\end{abstract}


\section{Introduction}
Boolean functions are fundamental in several symmetric cryptography applications. They are widely used to design \emph{Substitution Boxes} (S-Boxes) for block ciphers, and in certain types of stream ciphers such as the \emph{combiner model} and the \emph{filter model}~\cite{{Carlet21}}. To withstand cryptanalytic attacks, the Boolean functions adopted in these ciphers have to satisfy a number of cryptographic properties, some of which include \emph{balancedness}, high \emph{nonlinearity} and \emph{algebraic degree}, low
\emph{absolute indicator}, \emph{correlation immunity} and \emph{propagation criterion}.

Most of these properties cannot be satisfied simultaneously, since they induce several theoretical bounds and constraints among them. On the other hand, an exhaustive exploration to find the Boolean functions of $n$ variables achieving the best trade-off concerning a specific set of properties is not feasible in general, since the cardinality of the corresponding search space is $2^{2^n}$. As a consequence, the optimization of the cryptographic properties of Boolean functions is an important open problem in the design of symmetric ciphers.

Several heuristic techniques have been developed in the literature to discover Boolean functions satisfying good combinations of cryptographic properties, including Genetic Algorithms (GA)~\cite{Millan98,Manzoni20}, Simulated Annealing (SA)~\cite{Clark02} and Genetic Programming~\cite{Picek13,Picek16}.

The aim of this paper is to investigate the application of Particle Swarm Optimization (PSO) to the search of balanced Boolean functions with good cryptographic properties. Our starting point is the discrete variant of the PSO algorithm proposed by Hu et al. in~\cite{Hu03}. There, the authors considered combinatorial optimization problems over the Hamming cube, where the position of a particle in the swarm is represented by a $n$-dimensional binary vector. The velocity vector encodes for each coordinate the \emph{probability} through which a particle flips its corresponding binary value.

Since a Boolean function is uniquely identified by the binary vector of its truth table, one could directly apply Hu et al. discrete PSO to search for Boolean functions with good cryptographic properties. However, a basic requirement for Boolean functions in cryptographic designs is their balancedness, i.e. the property that their truth table vectors are composed of an equal number of zeros and ones. We thus modify the discrete PSO algorithm of~\cite{Hu03} and constrain it to search only over the space of balanced Boolean functions, thus removing this criterion from the optimization of the fitness function.

The main contributions of this paper are summarized as follows:

\begin{itemize}
\item We implement a new update method for the positions of the particles in a discrete PSO algorithms, which preserves the Hamming weights of the truth tables. This operator uses the velocity (i.e., the probability) of a particle over a specific coordinate to determine whether to \emph{swap} the corresponding bit with another one randomly chosen in the truth table.
\item We integrate the Hill Climbing procedure described in~\cite{Millan98}, to enhance the nonlinearity and the deviation from correlation immunity of the Boolean functions searched by our modified discrete PSO.
\item We perform parameter tuning for the social and cognitive constants, inertia, and maximum velocity parameters in the PSO velocity equation by employing two meta-optimization techniques: Local Unimodal Sampling (LUS) and Continuous Genetic Algorithms (CGA). While LUS has been used in previous research to fine-tune PSO parameters~\cite{Pedersen10}, to the best of our knowledge, CGA has not been applied to this meta-optimization task. Our results show that CGA achieves better results.
\item We finally employ the parameters optimized through CGA to run the PSO algorithm on the spaces of Boolean functions $n$ variables with $7 \le n \le 12$. The results of the experiments are reported and compared with those achieved by other heuristic methods published in the literature, focusing only on the best solutions found. We
observe that our PSO algorithm is able to find Boolean functions with similar or better combinations of nonlinearity, correlation immunity and propagation criterion than the ones produced by other methods, especially when the number of variables is less than $10$. It is also found that the properties (especially the nonlinearity) get worse as the number of variables increases, suggesting that further parameters tuning is required.
\end{itemize}

This paper is an extended version of our previous poster published in GECCO 2015~\cite{Mariot15}. In this sense, the present manuscript gives a full account of our discrete PSO algorithms and provides a complete overview of the performed experiments.

The rest of this paper is organized as follows. Section~\ref{sec:bg} recalls the background definitions and results about Boolean functions and their cryptographic properties. Section~\ref{sec:rel-work} gives an overview of the works in the literature pertaining the use of metaheuristics to optimize the cryptographic properties of Boolean functions. Section~\ref{sec:pso} describes our discrete PSO algorithm, introducing the weight-preserving operator for the position of the particles and defining the fitness functions used in our experimental evaluation. Section~\ref{sec:tuning} addresses the tuning of the PSO parameters using the LUS and CGA meta-optimizers. Section~\ref{sec:exp} describes the experimental campaign used to assess our discrete PSO algorithm, and compare the obtained results with those of other other optimization methods. Section~\ref{sec:outro} summarizes the main contributions of the paper, and discusses several directions for future research.

\section{Basics of Boolean Functions}
\label{sec:bg}
This section provides an overview of the essential concepts related to Boolean functions that will be referenced throughout this paper. While our discussion here focuses solely on cryptographic properties relevant to our optimization problem, for a more extensive exploration of this subject, readers are referred to Carlet's recent book~\cite{Carlet21}. As a standard notation, we denote $\F_2$ as the finite field comprising two elements and $\F_2^n$ as the $\F_2$-vector space consisting of binary $n$-tuples. In this context, vector addition represents the bitwise XOR, and scalar multiplication involves the logical AND operation between a single bit $a \in \F_2$ and each coordinate of a vector $x \in \F_2^n$. When referencing a binary vector $x$, $w_H(x)$ denotes its Hamming weight, that is, the number of non-zero coordinates within $x$.

\subsection{Representations of Boolean Functions}
\label{subsec:bool-rep}
A \emph{Boolean function} of $n$ variables is a mapping $f:\F_2^n \rightarrow \F_2$. When the elements of $\F_2^n$ are arranged in lexicographical order, the truth table of $f$, denoted as $a_f$, is a binary vector of length $2^n$. This truth table defines, for all inputs $x=(x_1,\cdots,x_n) \in \F_2^n$, the respective output value $f(x)$.

Another common representation of Boolean functions is the \emph{Algebraic Normal Form} (ANF). Given $f:\F_2^n \rightarrow \F_2$ and $x \in \F_2^n$, the ANF associated to $f$ is a multivariate polynomial $P_f(x)$ in the quotient ring $\F_q[x_1,\cdots,x_n]/(x_1^2 \oplus x_1, \cdots, x_n^2 \oplus x_n)$ defined as follows:
\begin{equation}
\label{eq:anf}
P_f(x) = \bigoplus_{I \subseteq [n]} a_I \cdot \left( \prod_{i \in I} x_i
\right) \enspace ,
\end{equation}
where $\oplus$ and $\cdot$ respectively denote sum and product over $\F_2$, and $[n]=\{1,\cdots,n\}$. For all $I \subseteq [n]$, the coefficient $a_I$ is uniquely determined by the \emph{M\"{o}bius transform}:
\begin{displaymath}
a_I = \bigoplus_{x \in \F_2^n: \mathcal{S}(x)\subseteq I} f(x) \enspace ,
\end{displaymath}
where $\mathcal{S}(x)=\{i \in [n]: x_i \ne 0\}$ is the \emph{support} of $x$.

Another representation of Boolean functions is the \emph{Walsh transform}, which is useful to characterize several cryptographic properties. Given a $n$-variable Boolean function $f:\F_2^n \rightarrow \F_2$, the Walsh transform $W_f: \F_2^n \rightarrow \Z$ of $f$ is defined for all $a \in F_2^n$ as
\begin{equation}
\label{eq:wht}
W_f(a) = \sum_{x \in \F_2^n} (-1)^{f(x)} \cdot (-1)^{a \cdot a} \enspace ,
\end{equation}
\noindent
where $a \cdot x = a_1\cdot x_1 \oplus \cdots \oplus a_n \cdot x_n$ is the \emph{scalar product between} $a$ and $x$. 

Finally, the \emph{autocorrelation function}  $A: \F_2^n \rightarrow \Z$ of a Boolean function $f: \F_2^n \rightarrow \F_2$ is defined for all $s \in \F_2^n$ as:
\begin{equation}
\label{eq:ac}
A(s) = \sum_{x \in \F_2^n} W_f(x)\cdot W_f(x\oplus s)
\end{equation}

\subsection{Cryptographic Properties of Boolean Functions}
\label{subsec:prop}
The first important cryptographic property which can be defined using the truth table representation is balancedness:

\begin{definition}
\label{def:bal}
A Boolean function $f: \F_2^n \rightarrow \F_2$ is \emph{balanced} if  $w_H(a_f)=2^{n-1}$, i.e. the truth table of $f$ is composed of an equal number of zeros and ones.
\end{definition}
Balancedness is an essential cryptographic criterion, since biases in the output distribution of a Boolean functions can be exploited for distinguishing attacks. Using the Walsh transform representation, remark that A Boolean function is balanced if and only if $W_f(\underbar{0}) = 0$, where $\underbar{0}$ is the null vector of $\F_2^n$.

A second essential criterion is the algebraic degree of a Boolean function, which is defined in terms of its ANF as follows:

\begin{definition}
\label{def:deg}
The \emph{algebraic degree} of a Boolean function $f:\F_2^n \rightarrow \F_2$ is the degree of the largest nonzero monomial in its ANF $P_f(x)$. Formally, $deg(f)$ is defined as
\begin{equation}
\label{eq:deg}
deg(f) = max\{|I|: I \subseteq [n], a_I \ne 0 \} \enspace .
\end{equation}
\end{definition}
Boolean functions having degree 1 are called \emph{affine functions}. The algebraic degree of Boolean functions should be as high as possible in order to resist attacks based on the \emph{Berlekamp-Massey algorithm} in stream ciphers~\cite{Massey69,Rueppel87} and higher order differential attacks in block ciphers~\cite{Knudsen94}.

The distance from the set of all affine functions is an important indicator of the strength of a Boolean function against attacks based on linear approximations of a cipher. This distance is aptly named the \emph{nonlinearity} of the function, and it can be defined through the Walsh transform as follows:

\begin{definition}
\label{def:nl}
The \emph{nonlinearity} $Nl(f)$ of a Boolean function $f:\F_2^n \rightarrow \F_2$ is the minimum Hamming distance of $f$ from the set of affine functions, and it is computed as follows:
\begin{equation}
\label{eq:nl}
Nl(f) = 2^{n-1} - \frac{1}{2} W_{max}(f) \enspace ,
\end{equation}
where $W_{max}(f) = \max_{a \in \F_2^n} \{|W_f(a)|\}$.
\end{definition}
Boolean functions having high nonlinearity provide better confusion. In particular, highly nonlinear Boolean functions should be used to resist fast correlation attacks in stream ciphers~\cite{Meier89} and linear cryptanalysis in block ciphers~\cite{Matsui93}.

It is known that, if the number of variables $n$ is even, the class of \emph{bent functions} reaches the maximum value of nonlinearity $2^{n-1}-2^{\frac{n-2}{2}}$. However, such functions are not balanced, thus they cannot be used directly in the design of symmetric cryptosystems. Determining the maximum nonlinearity for non-bent Boolean functions when $n$ is even, or for generic Boolean functions when $n$ is odd, is still an open problem for all $n>7$~\cite{Carlet21}.

A second important cryptographic criterion that can be defined with the Walsh transform is \emph{correlation immunity}:

\begin{definition}
\label{def:ci}
Given $k \in \{1,\cdots,n\}$, a Boolean function $f:\F_2^n \rightarrow \F_2$ is $k$-th order \emph{correlation immune} (denoted by $CI(k)$) if, by fixing the values of at most $k$ input coordinates, the truth tables of the corresponding restrictions of $f$ all have the same Hamming weight. This condition is verified if and only if $W_f(a) = 0$ for all $a \in \F_2^n$ such
that $1 \le w_H(a)\le k$ (see~\cite{XiaoM88}).
\end{definition}
A balanced Boolean function which is also $k$-th order correlation immune is called $k$-\emph{resilient}. Boolean functions used in stream ciphers based on the combiner model should be resilient of high order to resist correlation attacks~\cite{Siegenthaler84}. 

The maximum absolute value $AC_{max}$ for $s \in \F_2^n \setminus \{\underbar{0}\}$ of the autocorrelation function $A_f$ of a Boolean function $f: \F_2^ \to \F_2$ is called the \emph{absolute indicator} of $f$. This quantity should be as low as possible to withstand cube attacks in block ciphers~\cite{Dinur11}. Another cryptographic property related to the autocorrelation function is the \emph{propagation criterion}:

\begin{definition}
\label{def:pc}
Given $ l \in \{1,\cdots,n\}$, a Boolean function $f:\F_2^n \rightarrow \F_2$ satisfies the \emph{propagation criterion} $PC(l)$ if, for all nonzero vectors $s\in \F_2^n$ such that $w_H(s) \le l$, the function $f(x)\cdot f(x\oplus s)$ is balanced. This condition is met if and only if
$A(s) = 0$ for all $s \in \F_2^n$ such that $1 \le w_H(s) \le l$.
\end{definition}
The propagation criterion $PC(1)$ corresponds to the \emph{Strict Avalanche Criterion} (SAC) introduced in~\cite{Webster85}, which states that by complementing a single input coordinate $x_i$ the probability that the output of $f$ will change is $1/2$. Boolean functions in block ciphers should satisfy this property with a high order $l$ to differential cryptanalysis attacks~\cite{Biham90}.

Remark that most of the above criteria cannot be satisfied simultaneously. In particular, given a $k$-resilient, $PC(l)$ Boolean function of $n$ variables $f:\F_2^n \rightarrow \F_2$ having algebraic degree $deg(f)$ and nonlinearity $Nl(f)$, the following bounds hold:

\begin{itemize}
\item \emph{Siegenthaler's bound}~\cite{Siegenthaler84}: $deg(f) \le n - 1 - k$.
\item \emph{Sarkar-Maitra's bound}~\cite{Sarkar00}: $Nl(f) \le 2^{n-1} - 2^{k+1}$.
\item \emph{CI-PC bound}:~\cite{Charpin02}: $k+l \le n-1$.
\end{itemize}

\section{Related Work}
\label{sec:rel-work}
In this section, we give a brief overview of the literature related to the optimization of Boolean functions for cryptographic use through metaheuristics. For a more thorough review of the subject, we refer the reader to the recent survey paper by Djurasevic et al.~\cite{Djurasevic23}.

Millan et al.~\cite{Millan98} proposed a Genetic Algorithm (GA) coupled with Hill Climbing (HC) to evolve the truth tables of highly nonlinear balanced Boolean functions having low deviations from correlation immunity and propagation criterion. Later, Clark et al.~\cite{Clark02} devised a Simulated Annealing (SA) procedure to optimize the nonlinearity and the absolute indicator of Boolean functions as well as their correlation immunity and propagation criteria, which achieved better performances than GA. Aguirre et al. designed in~\cite{Aguirre07} a multi-objective Random Bit Climber which was able to generate more efficiently Boolean functions having good nonlinearity and absolute indicator. More recently, Genetic Programming (GP) has also been used by Picek et al.~\cite{Picek13} to evolve strong cryptographic Boolean functions of $8$ variables. The work has later been expanded in~\cite{Picek16}, where the authors compared four different evolutionary algorithms (namely GA, Evolutionary Strategies, GP and Cartesian GP) against three objective functions, each targeting a different subset of cryptographic properties. The results on Boolean functions of $n=8$ variables showed that GP (both in its basic and Cartesian variant) achieved the best results. More recently, Manzoni et al.~\cite{Manzoni20} performed a systematic investigation of balanced crossover operators in GA for the evolution of cryptographic Boolean functions. In particular, they compared Millan et al.'s counter-based crossover proposed in~\cite{Millan98} with other two balanced operators based on the map-of-ones and zero-length encodings. The idea is to restrict the search space explored by the GA by generating only balanced bitstrings during crossover and mutation, which then correspond to balanced Boolean functions. The experimental evaluation pointed out that the map-of-ones crossover was the best performing one.

The works above directly search the space of Boolean functions, using different representations (e.g. bitstrings encoding the truth table in GA or Boolean trees in GP). A different approach is to start from Walsh spectra that already encodes good properties (such as balancedness, correlation immunity and high nonlinearity) and then work backward by applying the inverse Walsh transform. The problem is that in general the resulting function will not be Boolean, but rather \emph{pseudo-Boolean}. The optimization objective thus becomes find a proper permutation of a suitable Walsh spectra that corresponds to an actual Boolean function. Clark et al.~\cite{Clark04} were the first to pioneer this spectral inversion method, designing a simulated annealing algorithm to optimize the Walsh spectra of pseudo-Boolean functions with good cryptographic properties. Building upon this approach, Mariot and Leporati~\cite{Mariot15b} later devised a genetic algorithm to evolve the spectra of pseudo-Boolean plateaued functions, remarking that it achieved better performances than simulated annealing for functions of $n=6$ variables.

A third approach is to evolve \emph{algebraic constructions} of Boolean functions instead of directly optimizing single solutions, be that in the direct search setting or the spectral inversion method. An algebraic construction takes in input some parameter (such as the number of variables and some existing Boolean functions with good cryptographic properties) and outputs new Boolean functions with similar good properties. Along this research line, Picek and Jakobovic~\cite{Picek16b} explored the use of GP to evolve algebraic constructions for bent functions. Similarly, Carlet et al.~\cite{Carlet21b} utilized GP to enhance Boolean functions derived from algebraic constructions. This lead to constructions for the so-called Hidden-Weight Boolean functions with an improved nonlinearity. Further, Mariot et al.~\cite{Mariot22} investigated an alternative algebraic construction based on cellular automata (CA), and designed an evolutionary strategies algorithm to find CA local rules that induce bent and semi-bent functions under such construction. Finally, more recently Carlet et al.~\cite{Carlet22} applied GP to evolve algebraic constructions for balanced functions with high nonlinearity. Interestingly, their results showed that most of the constructions obtained by GP actually corresponds to the well-known direct sum construction, which was already discovered by mathematicians several decades earlier.

\section{PSO Algorithm}
In this section, we first recall the basic concepts of Particle Swarm Optimization and then introduce its discrete variant proposed by Kennedy and Eberhart. Next, we describe our method to update the position of a particle while retaining its balancedness (therefore restricting the search space explored by the discrete PSO). Further, we define the fitness functions that target specific cryptographic properties of Boolean functions, and finally we describe the overall procedure of our discrete Particle Swarm Optimizer.

\label{sec:pso}
\subsection{Overview of Discrete PSO}
\label{subsec:dpso}
\emph{Particle Swarm Optimization} (PSO) is a metaheuristic optimization method initially introduced by Kennedy and Eberhart~\cite{Kennedy95}. PSO fundamentally operates by encoding a collection of potential solutions of an optimization problem as a \emph{swarm of particles}. These particles move collectively within a specified search space, typically a subset of $\mathbb{R}^m$. At each iteration $t \in \N$, the present \emph{position} of the $i$-th particle $x_i^{(t)}\in \mathbb{R}^m$ gets updated through the following equation:
\begin{displaymath}
x_i^{(t+1)} = x_i^{(t)} + v_i^{(t)} \enspace ,
\end{displaymath}
where $v_i^{(t)} \in \mathbb{R}^m$ denotes the \emph{velocity vector} of the $i$-th particle at time $t$. The solution encoded in the new position of the particle is then evaluated against a \emph{fitness function}, which is usually the objective function to be optimized. Each coordinate $j \in \{1,\cdots,m\}$ of the $i$-th particle velocity is in turn stochastically updated as follows:
\begin{displaymath}
v_{ij}^{(t+1)} = w \cdot v_{ij}^{(t)} + R_{ij} \cdot \varphi \cdot (g_j - x_{ij}^{(t)}) + R_{ij} \cdot \psi \cdot (b_{ij} - x_{ij}^{(t)}) \enspace ,
\end{displaymath}
\noindent
in which $v_{ij}^{(t)}$, the velocity of the $i$-th particle along dimension $j$ during the current time step $t$, is weighted by the \emph{inertia} parameter $w \in \mathbb{R}$. On the other hand, the random value $R_{ij} \in [0,1]$ is a parameter sampled with uniform probability, while $\varphi$ and $\psi$ are constants respectively used to weight the influence of the \emph{global best} solution $g \in \mathbb{R}^m$ found so far in the neighborhood of the particle, and of the \emph{local best} solution $b_i\in \mathbb{R}^m$ found so far by the $i$-th particle itself. To keep the velocity of the particle in check, the parameter $v_{max}$ is also used to clip the value of each coordinate $v_{ij}^{(t+1)}$ of the velocity vector. Concerning the neighborhood of a particle, one can consider different shapes, such as the \emph{Von Neumann} topology and the \emph{ring} topology. For the remainder of this work, we will consider only on the \emph{fully informed particle} strategy~\cite{Mendes04}, where the global best simply amounts to the best solution found by the entire swarm of particles.

The PSO metaheuristic was successfully employed in addressing numerous continuous optimization problems, as extensively discussed in surveys like Poli et al.'s work~\cite{Poli08}. However, its application to discrete search spaces is not straightforward. To this end, Kennedy and Eberhart introduced a modification to their original PSO algorithm in~\cite{Kennedy97}, aiming to tackle binary optimization problems. In this variant, solutions are represented as $m$-bit vectors, mapped onto the $m$-dimensional hypercube $\F_2^m$. Consequently, particles traverse the vertices of this hypercube, where the velocity vector transforms into a \emph{probability vector}. For each coordinate $j\in {1,\cdots,m}$ and the $i$-th particle in the swarm, the position $x_i$ concerning dimension $j$ gets updated by sampling a Bernoulli random variable using parameter $p_{ij}$. Specifically, if a randomly sampled number $r \in [0,1]$ is smaller than $p_{ij}$, the $j$-th coordinate of $x_i$ updates to $1$; otherwise, it updates to $0$.

The main advantage of this discrete PSO version lies is that the same velocity equation of the basic PSO heuristic can be used for updating the particles' probability vectors, granted that their components are normalized within the interval $[0,1]$ to get meaningful probability values. To achieve this, Kennedy and Eberhart adopted the \emph{logistic function} in~\cite{Kennedy97}, defined for all $x \in \mathbb{R}$ as:
\begin{displaymath}
S(x) = \frac{1}{1+\exp{(-x)}} \enspace .
\end{displaymath}

\subsection{Position Update for Balanced Functions}
\label{subsec:pos-upd}
The PSO heuristic, detailed in the preceding section, can be directly applied to the optimization challenge of identifying Boolean functions of $n$ variables with robust cryptographic properties, utilizing the truth table representation. In this scenario, particles navigate within a space comprising $m=2^n$ binary vectors. However, the approach proposed by Kennedy and Eberhart in~\cite{Kennedy97} for updating the positions of the particles lacks control over their Hamming weights. Due to the independent sampling of each component in the probability vector, there is no assurance that the generated truth tables will maintain balance, a critical criterion for cryptographic Boolean functions. One potential solution, as explored in~\cite{Picek13} for Genetic Algorithms and Genetic Programming, involves incorporating an \emph{unbalancedness} penalty within the fitness function. Yet, our initial experiments revealed this method is inadequate with PSO, resulting in a notably low proportion of generated balanced functions. Therefore, it becomes necessary to employ an update operator that confines the search space to the realm of balanced Boolean functions.

Hu, Eberhart, and Shi~\cite{Hu03} adapted the discrete PSO algorithm to tackle \emph{permutation problems}. Their approach involves stochastic \emph{swapping} of values within the permutation vector, denoting the particle's position. Specifically, the $i$-th particle's $x_{ij}$ component undergoes a change, with a probability of $p_{ij}$, by swapping it with $x_{ik}$, where $k$ corresponds to $x_{ik} = g_j$. This adjustment aims to align the permutation represented by the vector $x_i$ with the global best solution $g$.

\begin{figure}[t]
\centering
\begin{tikzpicture}
[->,auto,node distance=1.5cm, empt node/.style={inner
    sep=0pt,outer sep=0pt}, rect
  node/.style={rectangle,draw,minimum width=0.6cm,minimum
    height=0.6cm, inner sep=0pt, outer sep=0pt},white
  node/.style={rectangle,draw=white,fill=white,minimum width=0.2cm,minimum
    height=0.6cm, inner sep=0pt, outer sep=0pt}]

    \node [rect node] (p1) { $0.3$};
    \node [empt node] (p1c) [above=0.1cm of p1] {$1$};
    \node [empt node] (p) [left=0.1cm of p1] {$v=$};
    \node [rect node] (p2) [right=0cm of p1] {$0.1$};
    \node [empt node] (p2c) [above=0.1cm of p2] {$2$};
    \node [rect node] (p3) [right=0cm of p2, minimum width=1.2cm] {$\cdots$};
    \node [empt node] (p3c) [above=0.1cm of p3] {$\cdots$};
    \node [rect node] (p5) [right=0cm of p3] {$0.8$};
    \node [empt node] (p5c) [above=0.1cm of p5] {$i$};
    \node [empt node] (p5a) [below=0.25cm of p5] {};
    \node [empt node] (p5b) [right=0.1cm of p5a] {sample $r\simeq U(0,1)$};
    \node [rect node] (p6) [right=0cm of p5, minimum width=1.2cm] {$\cdots$};
    \node [empt node] (p6c) [above=0.1cm of p6] {$\cdots$};
    \node [rect node] (p7) [right=0cm of p6] {$0.5$};
    \node [empt node] (p7c) [above=0.1cm of p7] {$j$};
    \node [rect node] (p8) [right=0cm of p7, minimum width=1.2cm] {$\cdots$};
    \node [empt node] (p8c) [above=0.1cm of p8] {$\cdots$};
    \node [rect node] (p9) [right=0cm of p8] {$0.3$};
    \node [empt node] (p9c) [above=0.1cm of p9] {$2^n$};

    \node [rect node] (g1) [below=0.5cm of p1] {$0$};
    \node [empt node] (g) [left=0.1cm of g1] {$gb=$};
    \node [rect node] (g2) [right=0cm of g1] {$1$};
    \node [rect node] (g3) [right=0cm of g2, minimum width=1.2cm] {$\cdots$};
    \node [rect node] (g5) [right=0cm of g3] {$1$};
    \node [empt node] (n5a) [below=0.25cm of g5] {};
    \node [empt node] (n5b) [right=0.1cm of n5a] {if $x_i \neq gb_i
        \Rightarrow$ apply swap};
    \node [rect node] (g6) [right=0cm of g5, minimum width=1.2cm] {$\cdots$};
    \node [rect node] (g7) [right=0cm of g6] {$0$};
    \node [rect node] (g8) [right=0cm of g7, minimum width=1.2cm] {$\cdots$};
    \node [rect node] (g9) [right=0cm of g8] {$1$};

    \node [rect node] (x1) [below=0.5cm of g1] {$0$};
    \node [empt node] (x) [left=0.1cm of x1] {$x=$};
    \node [rect node] (x2) [right=0cm of x1] {$1$};
    \node [rect node] (x3) [right=0cm of x2, minimum width=1.2cm] {$\cdots$};
    \node [rect node] (x5) [right=0cm of x3] {$0$};
    \node [rect node] (x6) [right=0cm of x5, minimum width=1.2cm] {$\cdots$};
    \node [rect node] (x7) [right=0cm of x6] {$1$};
    \node [empt node] (x6a) [below=0.3cm of x7] {};
    \node [empt node] (x7a) [below=0.4cm of x6] {swap $x_i,x_j$ s.t. $x_i\neq x_j$ AND $x_j \neq gb_j$};
    \node [rect node] (x8) [right=0cm of x7, minimum width=1.2cm] {$\cdots$};
    \node [rect node] (x9) [right=0cm of x8] {$0$};

    \draw[->] (p5) -- (g5);
    \draw[<->] (g5) -- (x5);
    \draw[->] (x6a) -| (x5.south);
    \draw[->] (x6a) -- (x7.south);
\end{tikzpicture}
\caption{Example of application of swap-based position update.}
\label{fig:ex-swap-up}
\end{figure}
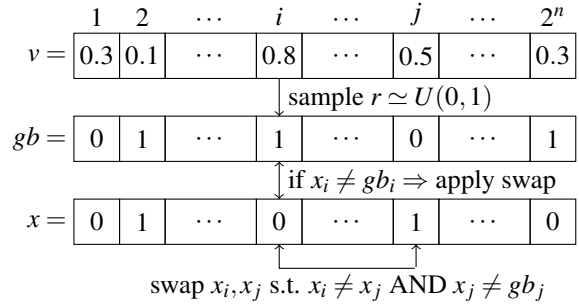

From a combinatorial standpoint, the collection of balanced Boolean functions with $n$ variables is isomorphic to the set of $\binom{2^{n}}{2^{n-1}}$ combinations. A subset of $2^{n-1}$ out of $2^{n}$ objects can be defined through its \emph{characteristic function}, essentially a balanced binary vector $x \in \F_2^m$, where $m=2^n$. Building on this observation, we extended the update operator proposed by Hu, Eberhart, and Shi to the domain of balanced combinations. Given the balanced binary vector $x_i\in\F_2^m$ and its corresponding probability vector $p_i\in [0,1]^m$, for each coordinate $j \in { 1, \cdots, m }$, a random number $r \in [0,1]$ is uniformly sampled. If $r$ is less than $p_{ij}$, a swap occurs as follows. Initially, the values of $x_{ij}$ and the global best at the same index, $g_j$, are compared. If these values match, no action is taken. Otherwise, a bit swap between $x_{ij}$ and another bit $x_{ik}$ occurs, where $k \ne j$ satisfies the conditions that $x_{ik}\neq g_k$ and $x_{ik} \neq x_{ij}$. This ensures the Hamming weight preservation and a decrease in Hamming distance from the global best solution by $2$. As multiple indices $k$ may meet these conditions, our update operator randomly selects one. Figure~\ref{fig:ex-swap-up} shows how the position update operator works on a specific example. 

This entire update process is reiterated utilizing the local best $b_i$ instead of the global best $g$. Consequently, $x_i$ undergoes modifications considering both the social attraction of the entire swarm and the cognitive attraction of the particle. Furthermore, if the particle's current position equals $g$, a random pair of bits in $x_i$ is swapped to prevent premature convergence, mirroring a solution proposed in~\cite{Hu03}.

The general pseudocode depicted in Algorithm~\ref{alg:upd-pso} implements our position update operator. The input parameters $x_i$ and $y$ represent balanced binary vectors, signifying the $i$-th particle's position within the swarm and either the global best $g$ or local best $b_i$, respectively. The length of $x_i$, denoted as $m=2^n$, is assumed to be unequal to $y$. The vector $p_i$ corresponds to the probability vector associated with the $i$-th particle. The subroutine {\sc Rand-Unif}() generates a random number $r \in [0,1]$ with a uniform distribution, essential for determining whether a swap is necessary by comparing it to $p_{ij}$. Additionally, the omitted subroutine {\sc Find-Cand-Swap}() is responsible for searching a suitable index for the swap, returning $0$ if no index is found.
\begin{algorithm}
\floatname{algorithm}{Algorithm}
\caption{\sc{Update-Bal-Pos}($x_i$, $y$, $p_i$, $m$)}
\label{alg:upd-pso}
\begin{algorithmic}

\FOR{$j$ := $1$ to $m$}
	\STATE $r$ := {\sc Rand-Unif}()
	\IF{($r$ < $p_{ij}$ AND $x_{ij} \ne y_j$)}
			\STATE $k$ := {\sc Find-Cand-Swap}($x_i$, $j$)
                        \IF{($k\ne 0$)}
			\STATE  Swap $x_{ij}$ with $x_{ik}$
                        \ENDIF
	\ENDIF
\ENDFOR
\end{algorithmic}
\end{algorithm}

\subsection{Fitness Functions}
\label{subsec:fitness}
To lay the groundwork for appropriate fitness functions, we first define the concept of \emph{deviation from correlation immunity}, originally introduced in~\cite{Millan98}:

\begin{definition}
The \emph{deviation from $k$-th order correlation immunity} of a Boolean function $f:\F_2^n \rightarrow \F_2$ is defined as
\begin{displaymath}
cidev_k(f) = max\{|W_f(a)|: a \in \F_2^n, \ 1 \le
w_h(a) \le k\} \enspace .
\end{displaymath}
\end{definition} 

Likewise, we also employ the following \emph{deviation from propagation criterion}:

\begin{definition}
Given $f:\F_2^n \rightarrow \F_2$, the \emph{deviation from propagation criterion} $PC(l)$ of $f$ is defined as
\begin{displaymath}
pcdev_l(f) = max\{|A(s)|: s \in \F_2^n, \ 1 \le
w_h(s) \le l\} \enspace .
\end{displaymath}
\end{definition}

We experimented with our Particle Swarm Optimizer using three fitness functions, all to be maximized. The initial function, $fit_1$, encompasses the three criteria of nonlinearity, deviation from first-order correlation immunity, and deviation from the Strict Avalanche Criterion, and it is defined as:
\begin{displaymath}
fit_1(f) = Nl(f) - \frac{cidev_1(f)}{4} - \frac{pcdev_1(f)}{8} \enspace .
\end{displaymath}
\noindent
As the Walsh and autocorrelation spectra of a balanced Boolean function are multiples of $4$ and $8$ respectively, the deviations in $fit_1$ are scaled by these factors. This fitness function bears resemblance to those outlined in~\cite{Millan98} for Genetic Algorithms. The referenced methods propose minimizing the \emph{normalized deviation} of the Boolean function, computed as the maximum value between $cidev_k(f)/4$ and $pcdev_l(f)/8$, or maximizing the difference between nonlinearity and $cidev_k(f)$. We adopted the latter approach as our second fitness function, setting $k=2$:
\begin{displaymath}
fit_2(f) = Nl(f) - cidev_2(f) \enspace .
\end{displaymath}
Finally, the third fitness function focuses on the nonlinearity and the absolute indicator of Boolean functions, two criteria that numerous heuristic methods presented in the literature~\cite{Clark02,Aguirre07,Picek13} have concurrently optimized together.
\begin{displaymath}
fit_3(f) = Nl(f) - AC_{max}(f) \enspace .
\end{displaymath}
Remark that none of the aforementioned fitness functions considers the algebraic degree, in contrast to those utilized in~\cite{Picek13}. This decision stems from two key reasons. First, the algebraic degree proves simpler to optimize compared to nonlinearity or correlation immunity. As $n$ tends toward infinity, a random Boolean function with $n$ variables tends to have an algebraic degree almost surely at $n-1$~\cite{{Carlet21}}. Hence, heuristic methods integrating algebraic degree in their fitness functions are likely to identify Boolean functions with maximum degree, although they might not fulfill $CI(k)$ or $PC(l)$. Second, as demonstrated in Section 5, our PSO algorithm uncovers Boolean functions achieving Siegenthaler's bound, even without considering algebraic degree within our fitness functions.

\subsection{Overall PSO Algorithm}
\label{subsec:pso-alg}
To enhance the efficiency of our Particle Swarm Optimizer, we integrated it with the Hill Climbing (HC) algorithm devised by Millan, Clark, and Dawson~\cite{Millan98}. This technique operates by exchanging a pair of bits in the truth table of a balanced Boolean function, aiming to elevate its nonlinearity and diminish its deviation from $CI(k)$. Hereafter, {\sc Nl-Ci(k)-Hc} denotes the HC procedure that elevates nonlinearity while decreasing $cidev_k(f)$, while {\sc Nl-Hc} refers to HC aimed solely at increasing nonlinearity. For an in-depth understanding of the general HC method, readers are directed to~\cite{Millan98}.

The type of Hill Climbing executed by our PSO algorithm hinges on the applied fitness function: for $fit_1$ and $fit_2$, we respectively employ {\sc Nl-Ci(1)-Hc} and {\sc Nl-Ci(2)-Hc}, whereas for $fit_3$, we utilize {\sc Nl-Hc}.

Here's an overview of the discrete Particle Swarm Optimizer's complete procedure:

\begin{enumerate}
    \item Initialize a swarm of size $N$. For each $i \in {1, \cdots, N}$, randomly generate a balanced binary vector $x_i \in \F_2^m$ and a probability vector $p_i \in [0,1]^m$, where $m=2^n$ and $n$ represents the Boolean functions' variable count.
    \item Compute the fitness value $fit_k$ of solution $x_i$ for each $k \in {1,2,3}$ and $i \in N$.
    \item Update the global best solution $g$ and the local best solutions $b_i$ for all $i \in {1,\cdots,N}$.
    \item Utilize the PSO velocity recurrence to update the probability vector $v_i$ for each $i \in {1,\cdots,N}$, followed by normalizing each coordinate using the logistic function.
    \item Update the position vector $x_i$ for each $i \in {1,\cdots,N}$. If $x_i=g$ or $x_i=b_i$, swap a random pair of bits in $x_i$. Otherwise, execute {\sc Update-Bal-Pos}($x_i$, $g$, $p_i$, $m$) and then {\sc Update-Bal-Pos}($x_i$, $b_i$, $p_i$, $m$).
    \item Apply the hill climbing optimization step {\sc Nl-Ci(k)-Hc} or {\sc Nl-Hc} as described in~\cite{Millan98} to all swarm particles based on the fitness function.
    \item If the maximum iteration count has been reached, output the global best solution $g$; otherwise, return to step 2.
\end{enumerate}

\section{Parameters Tuning}
\label{sec:tuning}
Extensive literature highlights the significant impact of the velocity parameters on PSO's performance~\cite{Shi98,Trelea03}. To this end, we adopted a systematic strategy using a meta-optimization approach.

Meta-optimization treats the selection of the governing parameters of an optimizer $O$ as an optimization problem in itself. An overarching meta-optimizer $M$ is applied to explore the parameter space, employing a meta-fitness function to evaluate $O$'s performance based on specific parameter combinations.

A feasible solution for the meta-optimization problem within our discrete PSO consists of a four-dimensional vector $(w,\varphi,\psi,v_{max}) \in \mathbb{R}^4$, defining the parameters in the velocity equation. To align with findings from~\cite{Kennedy97}, we constrained each parameter's value to the interval $[0,10]$. As for the overarching meta-optimizer selection, we opted to trial both \emph{Local Unimodal Sampling} (LUS) and \emph{Continuous Genetic Algorithms} (CGA, also known as \emph{Real-coded Genetic Algorithms}).

\subsection{Local Unimodal Sampling}
\label{subsec:lus}
LUS functions as a method of local exploration that continously tweaks the current solution $x$ by uniformly selecting a point $y$ within its defined neighborhood $N(x)$. In scenarios where maximization is the goal, if the fitness score of $y$ exceeds that of $x$, the solution transitions to $y$. Conversely, the size of $N(x)$ shrinks through a discount factor $\beta$ if $y$ does not outperform $x$. This mechanism aims to prevent premature convergence to a local optimal solution by consistently exploring within a neighborhood of a fixed size. The sampling process persists until it meets a termination condition, often indicated by a minimum threshold $\tau$ that sets the neighborhood size limit. Pedersen and Chipperfield~\cite{Pedersen10} utilized LUS to refine velocity parameters in a Particle Swarm Optimizer designed for training artificial neural network weights.

\subsection{Continuous Genetic Algorithms}
\label{subsec:cga}
CGA extends Genetic Algorithms to continuous optimization problems, representing a candidate solution's chromosome through a vector of real numbers instead of a binary string. As far as we know, CGA has not been previously utilized to fine-tune PSO parameters. Within our meta-optimization framework, we incorporated the \emph{flat operator}, introduced by Radcliffe~\cite{Radcliffe91}, as the crossover method for our CGA. For the mutation process, we relied on the straightforward \emph{random operator} proposed by Michalewicz~\cite{Michalewicz92}. The reproduction operator implements the \emph{roulette wheel method}, selecting individuals stochastically based on their fitness. Specifically, within a population of $P$ chromosomes, the next generation formation involves creating $P/2$ chromosome pairs using the roulette wheel method. Each pair $(x,y)$ generates an offspring of two chromosomes $(c_1,c_2)$ by applying the flat crossover operator with a probability of $p_c$ (if not applied, the chromosome pair $(x,y)$ remains unaltered). The random mutation operator is then applied with a probability of $p_m$ to each locus of the offspring chromosomes. Additionally, an \emph{elitist strategy} is used to retain the best individual in the succeeding generation.

\subsection{Meta-Fitness Function}
\label{subsec:meta-fit}
The \emph{meta-fitness function} used to drive the exploration for an optimal mix of PSO parameters constitutes the most computationally intensive phase. To gauge its performance under a specific set of parameters denoted by $x \in \mathbb{R}^4$, numerous iterations of our discrete PSO algorithm are necessary. In our examination, we focused on balanced Boolean functions defined on $n=7$ variables, employing a swarm comprising $N = 50$ particles evolved across $I = 100$ iterations. This algorithm underwent $R=30$ independent runs, each recording the fitness of the globally best solution $g$ at the final iteration.

Pedersen and Chipperfield \cite{Pedersen10} suggest using the average fitness value $\mu_g$, representative of the global best across all $R$ optimization runs, as the meta-fitness function. In our search for Boolean functions with suitable cryptographic criteria, we also considered the maximum fitness value $max_g$ achieved by the global best across these runs.

For Boolean functions of $7$ variables, the maximum attainable nonlinearity value is $Nl_{max}=56$ \cite{Patterson83}. Consequently, the highest value achievable by $max_g$ concerning fitness functions $fit_1$ and $fit_2$ is capped at $56$. Within the literature, no instance of a $7$-variable function has been reported with an absolute indicator $AC_{max}<16$, with conjectures proposing $16$ as the lowest plausible bound \cite{Zhang96}. Thus, the maximum value that $max_g$ can reach with respect to fitness function $fit_3$ under existing knowledge is $56-16=40$.

Therefore, our meta-fitness function can be defined for a parameter vector $x \in \mathbb{R}^4$ as:
\begin{displaymath}
mfit_k(x) = \mu_g + max_g \enspace ,
\end{displaymath}
\noindent
where $k\in \{1,2,3\}$ indicates the fitness function $fit_k$ which is being optimized by the PSO algorithm.

\subsection{Meta-Optimization Results}
\label{subsec:meta-res}
In line with the procedure outlined in~\cite{Pedersen10}, we conducted $M=6$ individual runs to evaluate the performance of both LUS and CGA, tallying up to a sum of $36$ meta-optimization experiments across various fitness functions, denoted as $fit_k$. Within the LUS domain, our approach involved setting the discount factor at $\beta=0.33$ and establishing a minimum threshold of neighborhood size at $\tau = 0.001$. Conversely, in employing the CGA meta-optimizer, we structured a population size of $P=20$ individuals, evolving over $G=100$ generations, with the crossover and mutation probabilities configured at $p_c=0.95$ and $p_m=0.05$ correspondingly.

Table~\ref{tab:meta-opt} compares the best parameters combinations found by LUS and CGA over the $6$ meta-optimization runs, for each fitness function $fit_k$.

\begin{table}[htbp]
\centering
\caption{Comparison of Best PSO Parameters}
\label{tab:meta-opt}
\begin{tabular}{cccccc}
\hline\noalign{\smallskip}
$fit_k$                & Method & $\mu_g$ & $\max_g$ & $mfit_k(f)$ \\ 
\noalign{\smallskip}\hline
\noalign{\smallskip}
\multirow{2}{*}{$fit_1$} &  LUS   &    52.7  &    56 &    108.7   \\
                         &  CGA   &    53    &    56 &    109     \\
\hline
\multirow{2}{*}{$fit_2$} &  LUS   &    46    &    52 &   98      \\
                         &  CGA   &    46.27 &    56 &  102.27  \\
\hline
\multirow{2}{*}{$fit_3$} &  LUS   &    30.87 &    40 &  70.86   \\
                         &  CGA   &    38.4  &    40 &  78.4    \\
\hline
\end{tabular}
\end{table}

It is evident that Continuous Genetic Algorithms (CGA) outperform Local Unimodal Sampling (LUS) across all three fitness functions. While there is only a slight difference in average fitness values $\mu_g$ for $fit_1$, the parameter configuration discovered by LUS for $fit_2$ did not enable the Particle Swarm Optimizer to attain the maximum fitness value of $56$. Additionally, for $fit_3$, the mean fitness $\mu_g$ achieved by LUS is significantly lower than that obtained by CGA. However, the better performance of CGA corresponds to a greater computational cost because Genetic Algorithms rely on a population-based heuristic. Specifically, in our experiments, a single CGA meta-optimization run demanded a considerable number of fitness evaluations, totaling close to 3.0 x $10^8$. This process took nearly 17 hours to complete on a 64-bit Linux machine with a Core i5 architecture, operating at 2.8 GHz. Conversely, with the selected $\beta$ and $\tau$ parameters, LUS conducted an average of 4971 fitness evaluations per meta-optimization run before reaching the minimum threshold. This approximately corresponds to around 1.3 hours of CPU time on the same machine.

\section{PSO Experiments}
\label{sec:exp}
\subsection{Experimental Setting}
\label{subsec:exp-set}
Now, we detail the experiments conducted using our Particle Swarm Optimizer. We selected the most optimal combination of velocity parameters evolved by the CGA meta-optimizer, as it achieved a higher meta-fitness value compared to those derived from LUS. The specific parameter values chosen for each fitness function are presented in Table~\ref{tab:cga-param}.

\begin{table}[htbp]
\centering
\caption{CGA-Evolved PSO Parameters}
\label{tab:cga-param}
\begin{tabular}{ccccc}
\hline\noalign{\smallskip}
$fit_k$ & $w$ & $\varphi$ & $\psi$ & $v_{max}$ \\ 
\noalign{\smallskip}\hline
\noalign{\smallskip}
$fit_1$ & 0.5067 & 2.8751 & 1.3587 & 3.5008 \\
$fit_2$ & 0.7614 & 2.0073 & 2.0273 & 2.7183 \\
$fit_3$ & 0.2828 & 2.1824 & 0.8951 & 4.2639 \\
\hline
\end{tabular}
\end{table}

Concerning the problem instances, we applied our PSO algorithm to search for balanced Boolean functions from $n=7$ to $n=12$ input variables. The number of particles and iterations were set respectively to $P=200$ and $I=400$. Finally, for each value of $n$ and fitness function $fit_k$, we carried out $R=100$ independent PSO runs.

\subsection{Best Solutions Found}
\label{subsec:best-sol}
Tables~\ref{tab:best-fit1} to~\ref{tab:best-fit3} show for each fitness function the cryptographic properties of the best balanced Boolean functions discovered by PSO, that is, the properties of the global best solution $g$ which scored the highest fitness value among all the $R=100$ optimization runs. We reported the algebraic degree as well, even if we did not adopt this criterion in any of the three fitness functions.

\begin{table}[htbp]
\centering
\caption{Best Boolean Functions Found, $fit_1$}
\begin{tabular}{ccccccc}
\hline\noalign{\smallskip}
Property  & 7  & 8   & 9   & 10  & 11  & 12   \\
\noalign{\smallskip}\hline
\noalign{\smallskip}
$Nl$      & 56 & 112 & 236 & 480 & 972 & 1972 \\
$deg$     & 5  & 6   & 7   & 8   & 9   & 10   \\
$cidev_1$ & 0  & 0   & 0   & 0   & 0   & 0    \\
$pcdev_1$ & 0  & 0   & 8   & 8   & 8   & 8    \\

\hline
\end{tabular}
\label{tab:best-fit1}
\end{table}

\begin{table}[htbp]
\centering
\caption{Best Boolean Functions Found, $fit_2$}
\begin{tabular}{ccccccc}
\hline\noalign{\smallskip}
Property  & 7  & 8   & 9    & 10  & 11  & 12   \\
\noalign{\smallskip}\hline
\noalign{\smallskip}
$Nl$       & 56 & 112 & 232 & 476 & 972 & 1972 \\
$deg$      & 4  & 6   & 7   & 8   & 9   & 10   \\
$cidev_1$  & 0  & 8   & 8   & 8   & 8   & 16   \\ 
$cidev_2$  & 0  & 8   & 8   & 8   & 8   & 16   \\
\hline
\end{tabular}
\label{tab:best-fit2}
\end{table}

\begin{table}[htbp]
\centering
\caption{Best Boolean Functions Found, $fit_3$}
\begin{tabular}{ccccccc}
\hline\noalign{\smallskip}
Property  & 7  & 8   & 9   & 10  & 11  & 12   \\
\noalign{\smallskip}\hline
\noalign{\smallskip}
$Nl$      & 56 & 116 & 236 & 480 & 976 & 1972 \\
$deg$     & 5  & 6   & 7   & 9   & 10   & 11   \\
$AC_{max}$ & 16 & 32  & 48  & 80  & 128   & 208 \\
\hline
\end{tabular}
\label{tab:best-fit3}
\end{table}
As a general observation, one can notice in Tables~\ref{tab:best-fit1} and~\ref{tab:best-fit2} that the Boolean functions discovered by PSO satisfying $CI(k)$ always have an algebraic degree of $n-1-k$, which is the maximum allowed by Siegenthaler's bound. Hence, these results empirically confirm that it is not necessary to consider the algebraic degree in the definition of the PSO fitness functions, as we mentioned in Section 3.4.

Looking in particular at Table~\ref{tab:best-fit1}, we can see that our PSO algorithm scales fairly well to higher numbers of variables with respect to the optimization of $cidev_1$, even if the CGA parameters were evolved only for the case $n=7$. As a matter of fact, all the best Boolean functions found by PSO
with $fit_1$ are first order correlation immune (and thus 1\nobreakdash-\hspace{0pt}resilient, since they are also balanced). Moreover, for $n=7$ and $n=8$ they also satisfy the Strict Avalanche Criterion $PC(1)$,
while for higher values of $n$ they reach the minimum deviation $pcdev_1 = 8$. Nevertheless, our Particle Swarm Optimizer is able to find Boolean functions of up to $n=11$ variables which satisfy both $CI(1)$ and $PC(1)$, even if their nonlinearity is lower (for a detailed comparison with other heuristic methods,
see Section 5.3).

On the other hand, Table~\ref{tab:best-fit2} shows that by using fitness function $fit_2$ the Particle Swarm Optimizer does not perform well when the number of variables is higher than $7$. In fact, $2$-resilient functions are found only for $n=7$, while in all other cases the deviation from $CI(2)$ is at
least $8$. However, it worths noting that the best solution of $7$ variables, besides satisfying with equality Siegenthaler's bound, achieves Tarannikov's bound on nonlinearity as well, since $56 = 2^{7-1} - 2^{2+1}$.

Finally, another different behaviour of the PSO algorithm can be observed using fitness function $fit_3$. Indeed, one can see from Table~\ref{tab:best-fit3} that as the number of variables grows the absolute indicator of the best solution gets worse. Nonetheless, for $n=8$ and $n=11$ the nonlinearity values
achieved with $fit_3$ are greater than those obtained using $fit_1$, while they are equal in all other cases. 

\subsection{Comparison with other Heuristics}
\label{subsec:comp}
We now compare the results of our Particle Swarm Optimizer with those obtained by other heuristic methods. Due to the great heterogeneity in the experimental settings and the parameters adopted in the relevant literature, a comprehensive comparison is not possible. For this reason, in Tables~\ref{tab:comp-ci1}
to~\ref{tab:comp-nl-ac} we summarise the results separately for each class of cryptographically significant balanced Boolean functions discovered by the PSO algorithm. A dash symbol in the tables indicates that the corresponding data is not available, either because the heuristic failed to discover Boolean functions
with those cryptographic properties or because that specific case was not considered for testing.

Table~\ref{tab:comp-ci1} reports the maximum nonlinearity achieved by $CI(1)$ functions. In this case, we used Genetic Algorithms (GA)~\cite{Millan98}, Directed Search Algorithm (DSA)~\cite{Pasalic99} and Simulated Annealing (SA)~\cite{Clark02} for the comparison. It can be seen that for $n=7$ variables our PSO algorithm manages to find $1$-resilient functions having maximum nonlinearity $56$, while SA stops at $52$. For $8 \le n \le 12$, the results achieved by PSO are globally similar to those of the other optimization methods, except in the case of $n=11$ variables where it reaches a maximum nonlinearity of $972$ instead of $976$. In particular, our PSO outperforms both Genetic Algorithms and Simulated Annealing for $n=9$ and $n=10$ variables.

\begin{table}[htbp]
\centering
\caption{Maximum Nonlinearity Achieved by $CI(1)$ Functions}
\begin{tabular}{ccccccc}
\hline\noalign{\smallskip}
Method               & 7  & 8   & 9   & 10  & 11   & 12   \\
\noalign{\smallskip}\hline
\noalign{\smallskip}
GA~\cite{Millan98}   & -  & 112 & 232 & 476 & 976  & 1972 \\ 
DSA~\cite{Pasalic99} & -  & 112 & 236 & 480 & 976  & -    \\
SA~\cite{Clark02}    & 52 & 112 & 232 & 476 & -    & -    \\
\hline
PSO                  & 56 & 112 & 236 & 480 & 972  & 1972 \\
\hline
\end{tabular}
\label{tab:comp-ci1}
\end{table}

In Table~\ref{tab:comp-ci1-pc1} the maximum nonlinearity of balanced Boolean functions which satisfy both $CI(1)$ and $PC(1)$ is considered. By comparing the results achieved by PSO and SA, we can see that also in this case the former reaches a higher value of nonlinearity for $n=7$ variables, while for $n=8$ it
is equal to SA. To our knowledge, no heuristic method has ever been applied to discover functions satisfying both $CI(1)$ and $PC(1)$ of $n>8$ variables. However, our PSO algorithm managed to find this kind of functions for up to $n=11$ variables, even though for $n>8$ they were not the best solutions
among all the optimization runs with respect to fitness function $fit_1$. The nonlinearity of these functions is reported in Table~\ref{tab:comp-ci1-pc1} as a reference for future research.

\begin{table}[htbp]
\centering
\caption{Maximum Nonlinearity Achieved by Functions satisfying both $CI(1)$ and
  $PC(1)$}
\begin{tabular}{ccccccc}
\hline\noalign{\smallskip}
Method          & 7  & 8   & 9   & 10  & 11   & 12   \\
\noalign{\smallskip}\hline
\noalign{\smallskip} 
SA~\cite{Clark02} & 52 & 112 & -   & -   &  -   & -    \\
\hline
PSO             & 56 & 112 & 232 & 476 &  968 & -    \\
\hline
\end{tabular}
\label{tab:comp-ci1-pc1}
\end{table}

Table~\ref{tab:comp-ci2} reports the maximum nonlinearity achieved by Boolean functions with minimal deviation from second order correlation immunity. In particular, the performances of PSO and GA are compared, since in this case we used the same fitness function defined in~\cite{Millan98}. As we already
discussed in Section 5.2, we can observe that our PSO algorithm does not generalise well to higher numbers of variables. As a matter of fact, PSO manages to reach the same results achieved by GA only for $n=8$ variables, while in all other cases either the nonlinearity or the deviation from $CI(2)$ is worse. We remark however that for $n=7$ the $2$-resilient functions found by PSO have the same value of nonlinearity as the ones discovered by SA in~\cite{Clark02}.

\begin{table}[htbp]
\centering
\caption{Comparison of $Nl$ and $cidev_2$ Values}
\begin{tabular}{cccccccc}
\hline\noalign{\smallskip}
Method                               &           & 7  & 8   & 9   & 10  & 11   & 12   \\
\noalign{\smallskip}\hline
\noalign{\smallskip}
\multirow{2}{*}{GA~\cite{Millan98}}  & $Nl$      & -  & 112 & 232 & 480 & 976  & 1972 \\
                                     & $cidev_2$ & -  & 4   & 8   & 8   &  8   & 8    \\ 
\hline
\multirow{2}{*}{PSO}                 & $Nl$      & 56 & 112 & 232 & 476 & 972  & 1972 \\
                                     & $cidev_2$ & 0  & 8   & 8   & 8   &  8   & 16   \\ 
\hline
\end{tabular}
\label{tab:comp-ci2}
\end{table}
\begin{table}[htbp]
\centering
\caption{Comparison of $Nl$ and $AC_{max}$ Values}
\begin{tabular}{cccccccc}
\hline\noalign{\smallskip}
Method                                &            & 7   & 8   & 9   & 10  & 11  &  12  \\
\noalign{\smallskip}\hline
\noalign{\smallskip}
\multirow{2}{*}{RBC~\cite{Aguirre07}}  & $Nl$       & 56  & 116 & -   & -   & -   & -    \\
                                      & $AC_{max}$ & 16  & 24  & -   & -   & -   & -    \\
\multirow{2}{*}{GP~\cite{Picek13}}     & $Nl$       & -   & 116 & -   & -   & -   & -    \\
                                      & $AC_{max}$ & -   & 32  & -   & -   & -   & -    \\
\multirow{2}{*}{SA~\cite{Clark02}}     & $Nl$       & 56  & 116 & 238 & 484 & 982 & 1986 \\
                                      & $AC_{max}$ & 16  & 24  & 40  & 56  & 88  & 128  \\ 
\hline
PSO                                   & $Nl$       & 56  & 116 & 236 & 480 & 976 & 1972 \\
                                      & $AC_{max}$  & 16  & 32  & 48  & 80  & 128 & 208  \\
\hline
\end{tabular}
\label{tab:comp-nl-ac}
\end{table}
Similar considerations can be made for the comparisons in Table~\ref{tab:comp-nl-ac}, which reports the maximum nonlinearity reached by Boolean functions having minimal absolute indicator. The benchmark heuristics in this case are Multi-Objective Random Bit Climber (RBC)~\cite{Aguirre07}, Genetic Programming (GP)~\cite{Picek13} and again SA. It can be observed that for $n=7$ variables PSO obtained the same results as RBC and SA, while for $n=8$ it discovered the same combination of $Nl$ and $AC_{max}$ featured by GP. However, for $n>8$ our PSO scored worse values than SA with respect to both nonlinearity and absolute indicator.

\section{Conclusions}
\label{sec:outro}
We presented a discrete PSO algorithm in our study to search for balanced Boolean functions spanning from $n=7$ to $n=12$ variables, focusing on robust cryptographic properties. Our experiments indicate that our PSO effectively generates Boolean functions exhibiting equivalent or superior combinations of nonlinearity, first-order correlation immunity, and the Strict Avalanche Criterion compared to other optimization methods. However, it demonstrates suboptimal performance when minimizing deviation from $CI(2)$ or the absolute indicator. This limitation might stem from the velocity parameters being evolved solely for $n=7$ variables, suggesting a need for further parameter refinement for $n\geq8$. Given the considerable computational expense of our meta-fitness function, employing the LUS meta-optimizer might be more preferable than CGA for this task.

Exploring future advancements in this realm presents several possibilities. One avenue is to examine the performance of our Particle Swarm Optimizer against other fitness functions, such as the one employed in~\cite{Clark02} for Simulated Annealing, which evaluates the flatness of a Boolean function's Walsh spectrum. Another interesting research direction involves refining the {\sc Update-Bal-Pos}() procedure to perform only swaps that enhance nonlinearity or reduce the deviation from $k$-th correlation immunity. This property could be obtained, for instance, by integrating the Hill Climbing algorithm within the update process, potentially offering new optimization capabilities.

\bibliographystyle{abbrv}
\bibliography{bibliography}

\end{document}